\documentclass[10pt,twocolumn,letterpaper]{article}

\usepackage{cvpr}
\usepackage{times}
\usepackage{epsfig}
\usepackage{graphicx}
\usepackage{amsmath}
\usepackage{amssymb}

\usepackage[utf8]{inputenc} 
\usepackage[T1]{fontenc}    
\usepackage{url}            
\usepackage{booktabs}       
\usepackage{amsfonts}       
\usepackage{nicefrac}       
\usepackage{microtype}      
\usepackage{algorithm}
\usepackage{algorithmic}
\usepackage{array}
\usepackage{bm}
\usepackage{multirow}

\def\obs{{\rm obs}}
\def\syn{{\rm syn}}

\usepackage[pagebackref=true,breaklinks=true,letterpaper=true,colorlinks,bookmarks=false]{hyperref}

 \cvprfinalcopy 

\def\cvprPaperID{3327} 

\ifcvprfinal\fi

\def\I{{\bf I}}
\def\B{{\bf B}}

\def\D{{\cal D}}
\def\S{{\cal S}}

\def\E{{\rm E}}

\def\tI{\tilde{\bf I}}

\def\tM{\tilde{M}}

\begin{document}

\title{Synthesizing Dynamic Patterns by Spatial-Temporal Generative ConvNet}

\author{Jianwen Xie, Song-Chun Zhu, and Ying Nian Wu\\
University of California, Los Angeles (UCLA), USA\\
{\tt\small jianwen@ucla.edu, sczhu@stat.ucla.edu, ywu@stat.ucla.edu}}

\maketitle

\begin{abstract}
Video sequences contain rich dynamic patterns, such as dynamic texture patterns that exhibit stationarity in the temporal domain, and action  patterns that are non-stationary in either spatial or temporal domain. We show that a spatial-temporal generative ConvNet can be used to model and synthesize  dynamic patterns. The model defines a probability distribution on the video sequence, and the log probability is defined by a spatial-temporal ConvNet that consists of multiple layers of spatial-temporal filters to capture spatial-temporal patterns of different scales. The model can be learned from the training video sequences by an ``analysis by synthesis'' learning algorithm that iterates the following two steps. Step 1 synthesizes video sequences from the currently learned model. Step 2 then updates the model parameters based on the difference between the synthesized video sequences and the observed training sequences. We show that the learning algorithm can synthesize realistic dynamic patterns. 
  \end{abstract}

\section{Introduction}

There are a wide variety of dynamic patterns in video sequences, including dynamic textures \cite{doretto2003dynamic} or textured motions \cite{wang2002generative} that  exhibit statistical stationarity or stochastic repetitiveness in the temporal dimension, and action  patterns that are non-stationary in either spatial or temporal domain. Synthesizing and analyzing such dynamic patterns has been an interesting problem. In this paper, we focus on the task of synthesizing dynamic patterns using a generative version of the convolutional neural network (ConvNet or CNN). 

The ConvNet \cite{lecun1998gradient, krizhevsky2012imagenet}  has proven to be an immensely successful discriminative learning machine. The convolution operation in the ConvNet is particularly suited for signals such as images, videos and sounds that exhibit translation invariance either in the spatial domain or the temporal domain or both. Recently, researchers have become increasingly interested in the generative aspects of ConvNet,  for the purpose of visualizing the knowledge learned by the ConvNet, or synthesizing realistic signals, or developing generative models that can be used for unsupervised learning. 

In terms of synthesis,  various approaches based on the ConvNet have been proposed to synthesize realistic static images \cite{Alexey2015, KarolICML2015, Denton2015a, Kulkarni2015, LuZhuWu2016}. However, there has not been much work in the literature on synthesizing dynamic patterns based on the ConvNet, and this is the focus of the present paper. 

Specifically, we propose to synthesize dynamic patterns by generalizing the generative ConvNet model recently proposed by \cite{XieLuICML}. The generative ConvNet can be derived from the discriminative ConvNet. It is a random field model or an energy-based model \cite{lecun2006tutorial, Ng2011} that is in the form of exponential tilting of a reference distribution such as the Gaussian white noise distribution or the uniform distribution. The exponential tilting is parametrized by a ConvNet that involves multiple layers of linear filters and rectified linear units (ReLU)  \cite{krizhevsky2012imagenet}, which seek to capture features or patterns at different scales. 

The generative ConvNet can be sampled by the Langevin dynamics. 
The model can be learned by the stochastic gradient algorithm  \cite{younes1999convergence}. It is an ``analysis by synthesis'' scheme that seeks to match the synthesized signals generated by the Langevin dynamics to the observed training signals.  Specifically, the learning algorithm iterates the following two steps after initializing the parameters and the synthesized signals. Step 1 updates the synthesized signals by the Langevin dynamics that samples from the currently learned model. Step 2 then updates the parameters based on the difference between the synthesized data and the observed data in order to shift the density of the model from the synthesized data towards the observed data. It is shown by  \cite{XieLuICML}  that the learning algorithm can synthesize realistic spatial image patterns such as textures and objects. 

In this article, we generalize the spatial generative ConvNet by adding the temporal dimension, so that the resulting ConvNet consists of multiple layers of spatial-temporal filters that seek to capture spatial-temporal patterns at various  scales. 
We show that the learning algorithm for training the spatial-temporal generative ConvNet can synthesize realistic dynamic patterns.  We also show that it is possible to learn the model from incomplete video sequences with either occluded pixels or missing frames, so that model learning and pattern completion can be accomplished simultaneously.

\section{Related work} 

Our work is a generalization of the generative ConvNet model of \cite{XieLuICML} by adding the temporal dimension. \cite{XieLuICML} did not work on dynamic patterns such as those in the video sequences. The spatial-temporal discriminative ConvNet was used by \cite{ji20133d} for analyzing video data. The connection between discriminative ConvNet and generative ConvNet was studied by \cite{XieLuICML}.

Dynamic textures or textured motions have been studied by \cite{doretto2003dynamic,wang2002generative,wang2004analysis, han2015video}. For instance,  \cite{doretto2003dynamic} proposed a vector auto-regressive model coupled with frame-wise dimension reduction by single value decomposition. It is a linear model with Gaussian innovations. \cite{wang2002generative} proposed a dynamic model based on sparse linear representation of frames. See \cite{you2016kernel} for a recent review of dynamic textures. The spatial-temporal generative ConvNet is a non-linear and non-Gaussian model and is expected to be more flexible in capturing complex spatial-temporal patterns in dynamic textures with multiple layers of non-linear spatial-temporal filters. 

Recently \cite{vondrick2016generating} generalized the generative adversarial networks \cite{goodfellow2014generative} to model dynamic patterns. Our model is an energy-based model and it also has an adversarial interpretation. See section \ref{sect:adv} for details. 

For temporal data, a popular model is the recurrent neural network \cite{williams1989learning, hochreiter1997long}. It is a causal model and it requires a starting frame. In contrast, our model is non-causal, and does not require a starting frame. Compared to the recurrent network, our model is more convenient and direct in capturing temporal patterns at multiple time scales.

\section{Spatial-temporal generative ConvNet}
\label{gen_inst}


\subsection{Spatial-temporal filters}

To fix notation, let $\I(x, t)$ be an image sequence of a video defined on the square (or rectangular) image domain $\D$ and the time domain ${\cal T}$, where $x = (x_1, x_2) \in \D$ indexes the coordinates of pixels, and $t \in {\cal T}$ indexes the frames in the video sequence. We can treat $\I(x, t)$ as a three dimensional function defined on $\D \times {\cal T}$.  For a spatial-temporal filter $F$,  we let $F*\I$ denote the filtered image sequence or feature map, and let $[F*\I](x, t)$ denote the filter response or feature at pixel $x$ and time $t$. 

The spatial-temporal ConvNet is a composition of multiple layers of linear filtering and ReLU non-linearity, as expressed by the following recursive formula: 
\begin{equation}
\begin{aligned}
 & [F^{(l)}_{k}  *\I] (x, t)  =    h\Bigg(\sum_{i=1}^{N_{l-1}}  \sum_{(y, s) \in \S_{l}} w^{(l, k)}_{i, y, s}    \\
 &  \times  [F^{(l-1)}_{i}*\I](x+y, t + s) + b_{l, k}\Bigg), 
\end{aligned}
\label{eq:ConvNet}
\end{equation}
where $l \in \{1, 2, ..., {L}\}$ indexes the layers.  $\{F^{(l)}_k, k = 1, ..., N_l\}$ are the filters at layer $l$, and $\{F^{(l-1)}_i, i = 1, ..., N_{l-1}\}$ are the filters at layer $l-1$.  $k$ and $i$ are used to index filters at layers $l$ and $l-1$ respectively, and $N_l$ and $N_{l-1}$ are the numbers of filters at layers $l$ and $l-1$ respectively. The filters are locally supported, so the range of $(y, s)$  is within a local support $\S_{l}$  (such as a $7 \times 7 \times 3$ box of image sequence). The weight parameters $(w^{(l, k)}_{i, y, s}, (y, s)  \in \S_l, i = 1, ..., N_{l-1})$ define a linear filter that operates on $(F^{(l-1)}_{i}*\I, i = 1, ..., N_{l-1})$. The linear filtering operation is followed by ReLU $h(r) = \max(0, r)$.  At the bottom layer, $[F^{(0)}_k*\I](x, t) = \I_k(x, t)$, where $k \in \{{\rm R, G, B}\}$ indexes the three color channels. Sub-sampling may be implemented so that in  $[F^{(l)}_{k}  *\I](x, t)$, $x \in \D_l \subset \D$, and $t \in {\cal T}_l \subset {\cal T}$.  

The spatial-temporal filters at multiple layers are expected to capture the spatial-temporal patterns at multiple scales. It is possible that the top-layer filters are fully connected in the spatial domain as well as the temporal domain (e.g., the feature maps are $1 \times 1$ in the spatial  domain) if the dynamic pattern does not exhibit spatial or temporal stationarity.

\subsection{Spatial-temporal generative ConvNet} 

The spatial-temporal generative ConvNet is an energy-based model or a random field model  defined on the image sequence $\I = (\I(x, t), x \in {\cal D}, t \in {\cal T})$. It is in the form of exponential tilting of a reference distribution $q(\I)$: 
\begin{equation}
p(\I; w) = \frac{1}{Z(w)} \exp \left[ f(\I; w)\right] q(\I), 
\label{eq:ConvNet-FRAME}
\end{equation}
where the scoring function $f(\I; w)$ is 
\begin{equation}
f(\I; w) =\sum_{k=1}^{K} \sum_{x \in {\cal D}_L} \sum_{t \in {\cal T}_L} [F_k^{(L)}*\I](x, t),  
\label{eq:scoring}
\end{equation}
where $w$ consists of all the weight and bias terms that define the filters $(F_k^{(L)}, k = 1, ..., K = N_L)$ at layer $L$, and $q$ is the Gaussian white noise model, i.e., 
\begin{equation}
q(\I) = \frac{1}{(2\pi\sigma^2)^{|{\cal D} \times {\cal T}|/2}} \exp\left[- \frac{1}{2\sigma^2} ||\I||^2\right], 
\label{eq:Gaussian}
\end{equation}   
where $|\D \times {\cal T}|$ counts the number of pixels in the domain $\D \times {\cal T}$. Without loss of generality, we shall assume $\sigma^2 = 1$. 

The scoring function $f(\I; w)$ in (\ref{eq:scoring}) tilts the Gaussian reference distribution into a non-Gaussian model. In fact, the purpose of $f(\I; w)$ is to identify the non-Gaussian spatial-temporal features or patterns. In the definition of $f(\I; w)$ in (\ref{eq:scoring}), we sum over the filter responses at the top layer $L$ over all the filters, positions and times. The spatial and temporal pooling reflects the fact that we assume the model is stationary in spatial and temporal domains. If the dynamic texture is non-stationary in the spatial or temporal domain, then  the top layer filters $F_k^{(L)}$ are fully connected in the spatial or temporal domain, e.g., ${\cal D}_L$ is $1 \times 1$. 

A simple but consequential property of the ReLU non-linearity is that $h(r) = \max(0, r) = 1(r>0) r$, where $1()$ is the indicator function, so that $1(r>0) = 1$ if $r>0$ and $0$ otherwise. As a result, the scoring function $f(\I; w)$ is piecewise linear \cite{montufar2014number}, and each linear piece is defined by the multiple layers of binary activation variables $\delta_{k, x, t}^{(l)}(\I; w) = 1\left([F_k^{(l)}*\I](x, t)>0\right)$, which tells us whether a local spatial-temporal pattern represented by the $k$-th filter at layer $l$,  $F_k^{(l)}$, is detected at position $x$ and time $t$. Let $\delta(\I; w) = \left(\delta_{k, x, t}^{(l)}(\I; w), \forall l, k, x, t \right)$ be the activation pattern of $\I$. Then $\delta(\I; w)$ divides the image space into a large number of pieces according to the value of $\delta(\I; w)$. On each piece of image space with fixed $\delta(\I; w)$, the scoring function $f(\I; w)$ is linear, i.e.,  
\begin{equation}
f(\I; w) = a_{w, \delta(\I; w)}  + \langle \I, B_{w, \delta(\I; w)}\rangle,
\end{equation}
 where both $a$ and $B$ are defined by $\delta(\I; w)$ and  $w$. In fact, $B = \partial f(\I; w)/\partial \I$, and can be computed by back-propagation, with $h'(r) = 1(r>0)$. The back-propagation process defines a top-down deconvolution process \cite{zeiler2011adaptive}, where the filters at multiple layers become the basis functions at those layers, and the activation variables at different layers in $\delta(\I; w)$ become the coefficients of the basis functions in the top-down deconvolution. 

$p(\I; w)$ in  (\ref{eq:ConvNet-FRAME})  is an energy-based model \cite{lecun2006tutorial, Ng2011}, whose energy function is a combination of the $\ell_2$ norm $\|\I\|^2$ that comes from the reference distribution $q(\I)$ and the piecewise linear scoring function $f(\I; w)$, i.e., 
\begin{equation}
\begin{aligned}
 {\cal E}(\I; w) &= -f(\I; w) + \frac{1}{2} \|\I\|^2\\
 & =\frac{1}{2}  \|\I\|^2 - \left(a_{w, \delta(\I; w)}  + \langle \I, B_{w, \delta(\I; w)}\rangle \right)\\
 &= \frac{1}{2} \|\I - B_{w, \delta(\I; w)}\|^2 + {\rm const}, 
\label{eq:ConvNet-FRAME3}
\end{aligned}
\end{equation}
where ${\rm const} = -a_{w, \delta(\I; w)} - \|B_{w, \delta(\I; w)}\|^2/2$, which is constant on the piece of image space with fixed $\delta(\I; w)$.

Since ${\cal E}(\I; w)$ is a piecewise quadratic function,  $p(\I; w)$ is piecewise Gaussian. On the piece of image space $\{\I: \delta(\I; w) = \delta\}$, where $\delta$ is a fixed value of $\delta(\I; w)$, $p(\I; w)$ is ${\rm N}(B_{w, \delta}, {\bf 1})$ truncated to  $\{\I: \delta(\I; w) = \delta\}$, where we use ${\bf 1}$ to denote the identity matrix. If the mean of this Gaussian piece,  $B_{w, \delta}$,  is within $\{\I: \delta(\I; w) = \delta\}$, then $B_{w, \delta}$ is also a local mode, and this local mode $\I$ satisfies a hierarchical auto-encoder, with a bottom-up encoding process $\delta = \delta(\I; w)$, and a top-down decoding process $\I = B_{w, \delta}$. In general, for an image sequence $\I$, $B_{w, \delta(\I; w)}$ can be considered a reconstruction of $\I$, and this reconstruction is exact if $\I$ is a local mode of ${\cal E}(\I; w)$. 

\subsection{Sampling and learning algorithm}

One can sample from $p(\I; w)$  of model (\ref{eq:ConvNet-FRAME}) by the Langevin dynamics: 
\begin{equation}
\I_{\tau+1} = \I_{\tau} - \frac{\epsilon^2}{2} \left[\I_\tau - \B_{w, \delta(\I_\tau; w)}\right] + \epsilon Z_\tau, \label{eq:Langevin}
\end{equation}
 where $\tau$ indexes the time steps,  $\epsilon$ is the step size, and  $Z_\tau \sim {\rm N}(0, {\bf 1})$. The dynamics is driven by the reconstruction error $\I -\B_{w, \delta(\I; w)}$. The finiteness of the step size $\epsilon$ can be corrected by a Metropolis-Hastings acceptance-rejection step. The Langevin dynamics can be extended to Hamiltonian Monte Carlo \cite{neal2011mcmc} or more sophisticated versions \cite{girolami2011riemann}.

The learning of $w$ from training image sequences $\{\I_m, m = 1, ..., M\}$ can be accomplished by the maximum likelihood. Let $L(w) = \sum_{m=1}^{M} \log p(\I; w)/M$, 
with $p(\I; w)$ defined in (\ref{eq:ConvNet-FRAME}), 
\begin{equation}
\frac{\partial L(w)}{\partial w} = \frac{1}{M} \sum_{m=1}^{M} \frac{\partial}{\partial w} f(\I_m; w) 
   -  \E_{w} \left[ \frac{\partial}{\partial w} f(\I; w) \right].
\end{equation}
The expectation can be approximated by the Monte Carlo samples \cite{younes1999convergence} produced by the Langevin dynamics. See Algorithm \ref{code:FRAME}  for a description of the learning and sampling algorithm. The algorithm keeps synthesizing image sequences from the current model, and updating the model parameters in order to match the synthesized image sequences to the observed image sequences. The learning algorithm keeps shifting the probability density or low energy regions of the model from the synthesized data towards the observed data.

\begin{algorithm}
\caption{Learning and sampling algorithm}
\label{code:FRAME}
\begin{algorithmic}[1]

\REQUIRE ~~\\
(1)  training image sequences $\{\I_m, m=1,...,M\}$ \\
(2) number of synthesized image sequences $\tilde{M}$\\
(3) number of Langevin steps $l$\\
(4) number of learning iterations $T$

\ENSURE~~\\
(1) estimated parameters $w$\\
(2) synthesized image sequences $\{\tI_m, m = 1, ..., \tilde{M}\}$ 

\item[]
\STATE Let $t\leftarrow 0$, initialize $w^{(0)}$.
\STATE Initialize $\tI_m$, for $m = 1, ..., \tilde{M}$. 
\REPEAT 
\STATE For each $m$, run $l$ steps of Langevin dynamics to update $\tI_m$, i.e., starting from the current $\tI_m$, each step 
follows equation (\ref{eq:Langevin}). 
\STATE Calculate  $H^{\obs} = \sum_{m=1}^{M} \frac{\partial}{\partial w} f(\I_m; w^{(t)})/M$, and
$H^{\syn} =  \sum_{m=1}^{\tM} \frac{\partial}{\partial w} f(\tI_m; w^{(t)})/\tM$.
\STATE Update $w^{(t+1)} \leftarrow w^{(t)} + \eta_t ( H^{\obs} - H^{\syn}) $,  with step size $\eta_t$. 
\STATE Let $t \leftarrow t+1$
\UNTIL $t = T$
\end{algorithmic}
\end{algorithm}

In the learning algorithm, the Langevin sampling step involves the computation of $\partial f(\I; w)/\partial \I$, and the parameter updating step involves the computation of $\partial f(\I; w)/\partial w$. Because of the ConvNet structure of $f(\I; w)$, both gradients can be computed efficiently by back-propagation, and the two gradients share most of their chain rule computations in back-propagation. 
In term of MCMC sampling, the Langevin dynamics samples from an evolving distribution because $w^{(t)}$ keeps changing. Thus the learning and sampling algorithm runs non-stationary chains. 

\subsection{Adversarial interpretation} \label{sect:adv}

Our model is an energy-based model 
\begin{align} 
p(\I; w) =\frac{1}{Z(w)}  \exp[-{\cal E}(\I; w)].
\end{align}
 The update of $w$ is based on $L'(w)$ which can be approximated by 
 \begin{align}
 \frac{1}{\tilde{M}} \sum_{m=1}^{\tilde{M}} \frac{\partial}{\partial w} {\cal E}(\tilde{\I}_m; w)
 - \frac{1}{M} \sum_{m=1}^{M} \frac{\partial}{\partial w} {\cal E}(\I_m; w), 
 \end{align}  
where $\{\tilde{\I}_m, m = 1, ..., \tilde{M}\}$ are the synthesized image sequences that are generated by the Langevin dynamics. At the zero temperature limit, the Langevin dynamics becomes gradient descent:
\begin{equation}
\tilde{\I}_{\tau+1} = \tilde{\I}_{\tau} - \frac{\epsilon^2}{2} \frac{\partial}{\partial \tilde{\I}} {\cal E}(\tilde{\I}_{\tau}; w). \label{eq:Langevin1}
\end{equation}
Consider the value function $V(\tilde{\I}_m, m = 1, ..., \tilde{M}; w)$: 
 \begin{align}
 \frac{1}{\tilde{M}} \sum_{m=1}^{\tilde{M}} {\cal E}(\tilde{\I}_m; w)
 -\frac{1}{M} \sum_{m=1}^{M}  {\cal E}(\I_m; w). 
 \end{align}  
The updating of $w$ is to increase $V$ by shifting the low energy regions from the synthesized image sequences $\{\tilde{\I}_m\}$ to the observed image sequences $\{\I_m\}$, whereas the updating of $\{\tilde{\I}_m, m = 1, ..., \tilde{M}\}$ is to decrease $V$ by moving the synthesized image sequences towards the low energy regions. This is an adversarial interpretation of the learning and sampling algorithm.
 It can also be considered a generalization of the herding method \cite{welling2009herding} from exponential family models to general energy-based models. 

In our work, we let $-{\cal E}(\I; w) = f(\I; w) - \|\I\|^2/2\sigma^2$. We can also let $-{\cal E}(\I; w) = f(\I; w)$ by assuming a uniform reference distribution $q(\I)$. Our experiments show that the model with the uniform $q$ can also synthesize realistic dynamic patterns.  

The generative adversarial learning  \cite{goodfellow2014generative, vondrick2016generating} has a generator network. Unlike our model which is based on a bottom-up ConvNet $f(\I; w)$, the generator network generates $\I$ by a top-down ConvNet $\I = g(X; \tilde{w})$ where $X$ is a latent vector that follows a known prior distribution, and $\tilde{w}$ collects the parameters of the top-down ConvNet. Recently \cite{HanLu2016} developed an alternating back-propagation algorithm to train the generator network, without involving an extra network. More recently, \cite{xie2016cooperative} developed a cooperative training method that recruits a generator network $g(X; \tilde{w})$ to reconstruct and regenerate the synthesized image sequences $\{\tilde{\I}_m\}$ to speed up MCMC sampling. 

\section{Experiments}

We learn the spatial-temporal generative ConvNet from video clips collected from DynTex++ dataset of \cite{ghanem2010maximum} and the Internet.  The code in the experiments is based on the MatConvNet of \cite{matconvnn} and MexConv3D of ~\cite{MexConv3D}.

We show the synthesis results by displaying the frames in the video sequences. We have posted the synthesis results on the project page \url{http://www.stat.ucla.edu/~jxie/STGConvNet/STGConvNet.html}, so that the reader can watch the videos.

\subsection{Experiment 1: Generating dynamic textures with both spatial and temporal stationarity}

\begin{figure}
\begin{center}

\includegraphics[width=.13\linewidth]{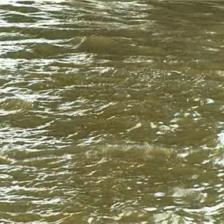}
\includegraphics[width=.13\linewidth]{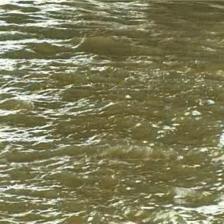}
\includegraphics[width=.13\linewidth]{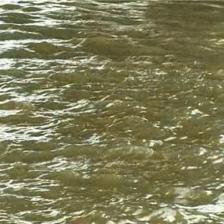}
\includegraphics[width=.13\linewidth]{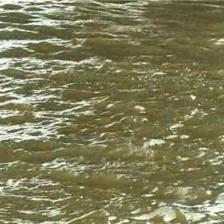}
\includegraphics[width=.13\linewidth]{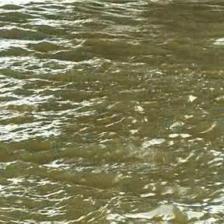}
\includegraphics[width=.13\linewidth]{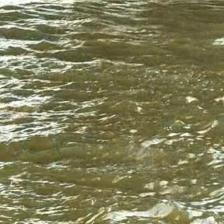}
\includegraphics[width=.13\linewidth]{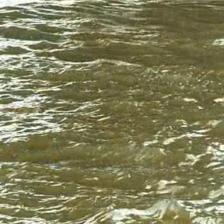}\\[3px]

\includegraphics[width=.13\linewidth]{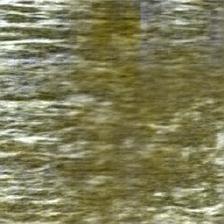}
\includegraphics[width=.13\linewidth]{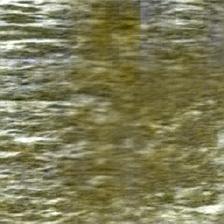}
\includegraphics[width=.13\linewidth]{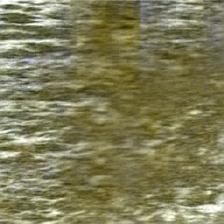}
\includegraphics[width=.13\linewidth]{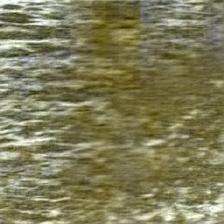}
\includegraphics[width=.13\linewidth]{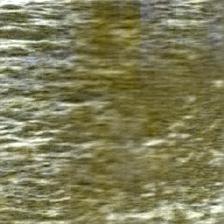}
\includegraphics[width=.13\linewidth]{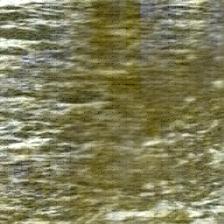}
\includegraphics[width=.13\linewidth]{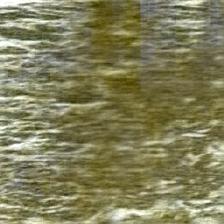} \\[3px]	

\includegraphics[width=.13\linewidth]{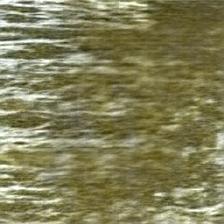}
\includegraphics[width=.13\linewidth]{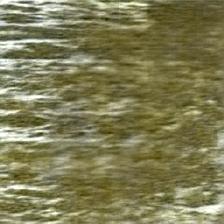}
\includegraphics[width=.13\linewidth]{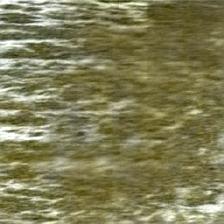}
\includegraphics[width=.13\linewidth]{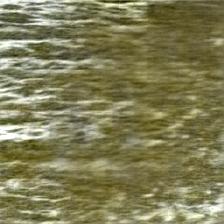}
\includegraphics[width=.13\linewidth]{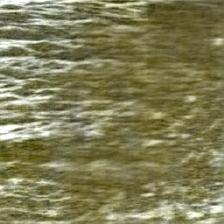}
\includegraphics[width=.13\linewidth]{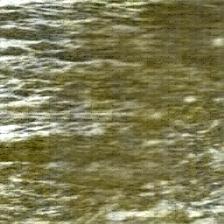}
\includegraphics[width=.13\linewidth]{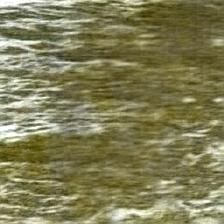}  \\ (a) river	\\[3px]

\includegraphics[width=.13\linewidth]{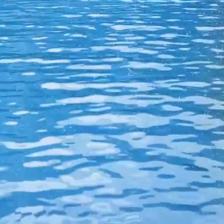}
\includegraphics[width=.13\linewidth]{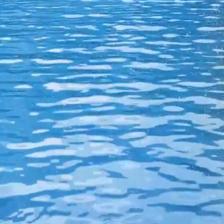}
\includegraphics[width=.13\linewidth]{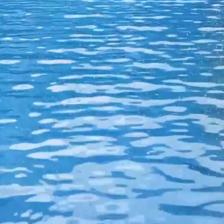}
\includegraphics[width=.13\linewidth]{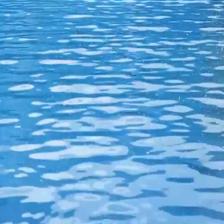}
\includegraphics[width=.13\linewidth]{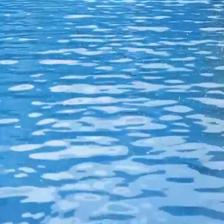}
\includegraphics[width=.13\linewidth]{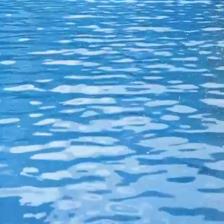}
\includegraphics[width=.13\linewidth]{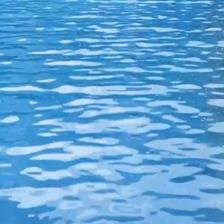}\\
[3px]

\includegraphics[width=.13\linewidth]{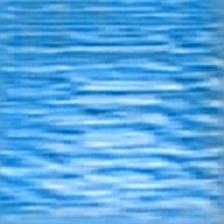}
\includegraphics[width=.13\linewidth]{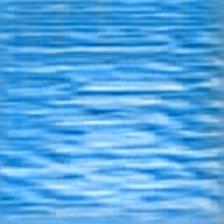}
\includegraphics[width=.13\linewidth]{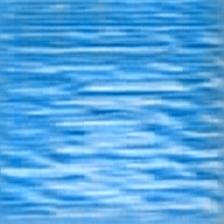}
\includegraphics[width=.13\linewidth]{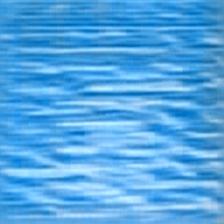}
\includegraphics[width=.13\linewidth]{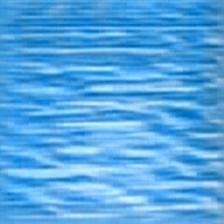}
\includegraphics[width=.13\linewidth]{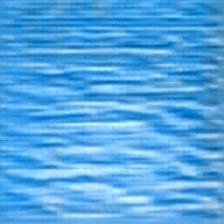}
\includegraphics[width=.13\linewidth]{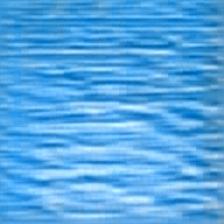}\\[3px]

\includegraphics[width=.13\linewidth]{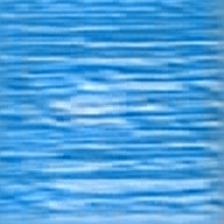}
\includegraphics[width=.13\linewidth]{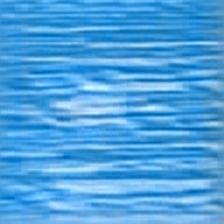}
\includegraphics[width=.13\linewidth]{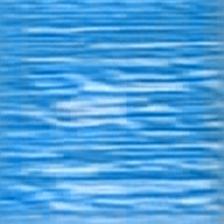}
\includegraphics[width=.13\linewidth]{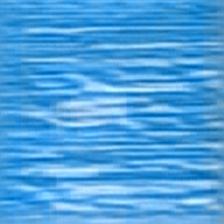}
\includegraphics[width=.13\linewidth]{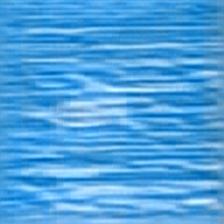}
\includegraphics[width=.13\linewidth]{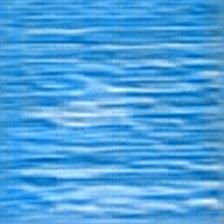}
\includegraphics[width=.13\linewidth]{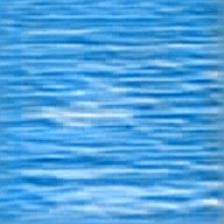}  \\ (b) ocean	\\[3px]	
	\caption{Synthesizing dynamic textures with both spatial and temporal stationarity. For each category, the first row displays the frames of the observed sequence, and the second and third  rows  display the corresponding frames of two synthesized sequences generated by the learning algorithm. (a) river. (b) ocean. }
	\label{fig:DTresults1}
\end{center}
\end{figure}

\begin{figure}
\begin{center}
\includegraphics[width=.15\linewidth]{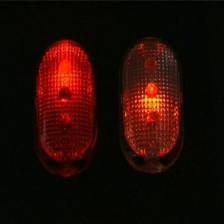}
\includegraphics[width=.15\linewidth]{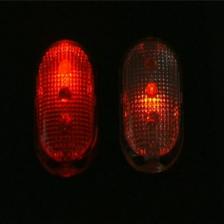}
\includegraphics[width=.15\linewidth]{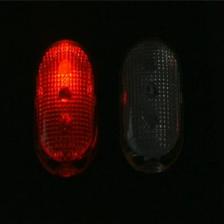}
\includegraphics[width=.15\linewidth]{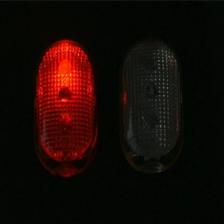} 
\includegraphics[width=.15\linewidth]{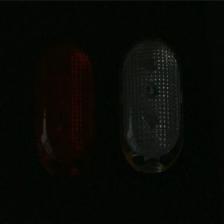} 
\includegraphics[width=.15\linewidth]{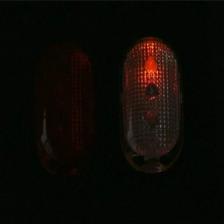}\\[3px]	

\includegraphics[width=.15\linewidth]{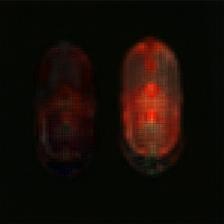}
\includegraphics[width=.15\linewidth]{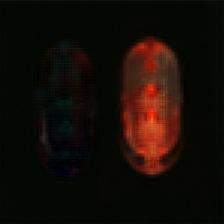}
\includegraphics[width=.15\linewidth]{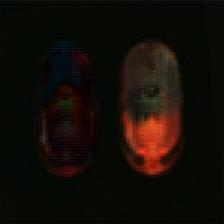}
\includegraphics[width=.15\linewidth]{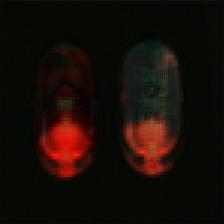} 
\includegraphics[width=.15\linewidth]{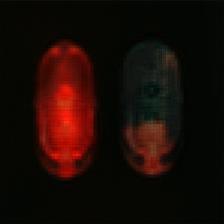} 
\includegraphics[width=.15\linewidth]{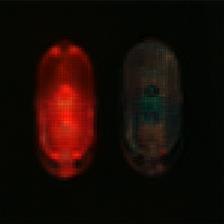}\\ (a) flashing lights \\[3px]

\includegraphics[width=.15\linewidth]{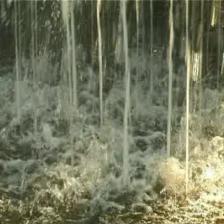}
\includegraphics[width=.15\linewidth]{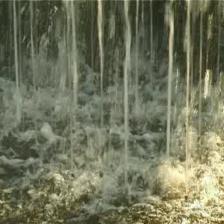}
\includegraphics[width=.15\linewidth]{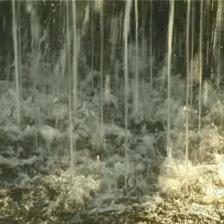}
\includegraphics[width=.15\linewidth]{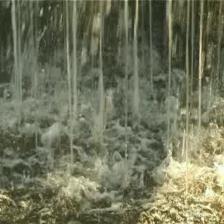}
\includegraphics[width=.15\linewidth]{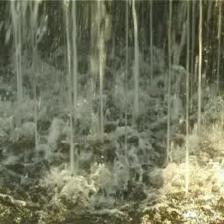} 
\includegraphics[width=.15\linewidth]{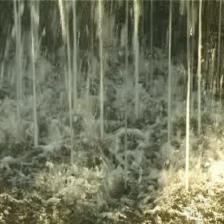} \\[3px]	

\includegraphics[width=.15\linewidth]{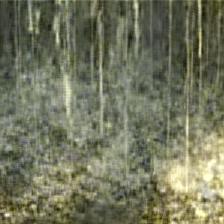}
\includegraphics[width=.15\linewidth]{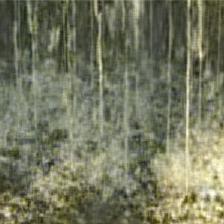}
\includegraphics[width=.15\linewidth]{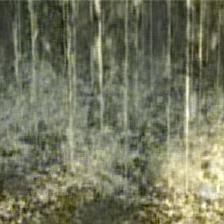}
\includegraphics[width=.15\linewidth]{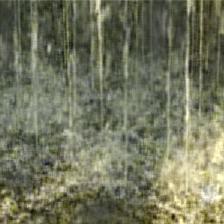}
\includegraphics[width=.15\linewidth]{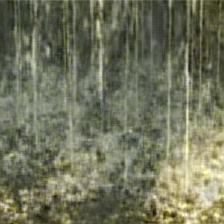} 
\includegraphics[width=.15\linewidth]{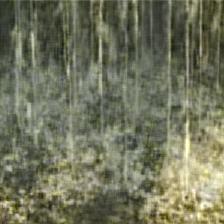} \\ (b) fountain\\[3px] 

\includegraphics[width=.15\linewidth]{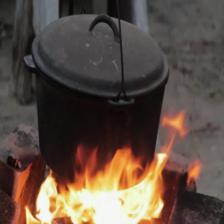}
\includegraphics[width=.15\linewidth]{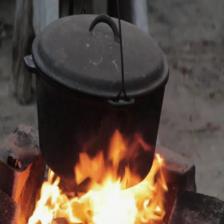}
\includegraphics[width=.15\linewidth]{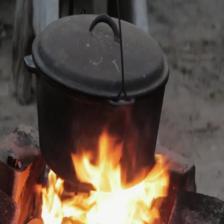}
\includegraphics[width=.15\linewidth]{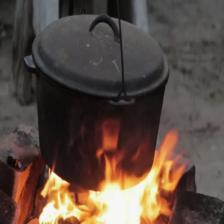}
\includegraphics[width=.15\linewidth]{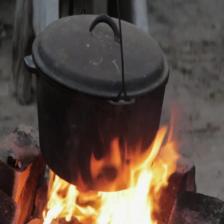} 
\includegraphics[width=.15\linewidth]{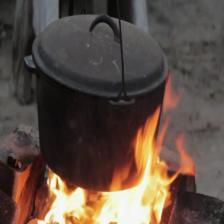}\\[3px]	 

\includegraphics[width=.15\linewidth]{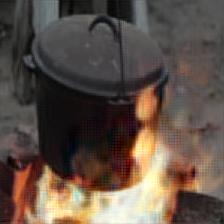}
\includegraphics[width=.15\linewidth]{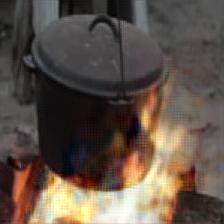}
\includegraphics[width=.15\linewidth]{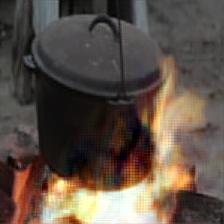}
\includegraphics[width=.15\linewidth]{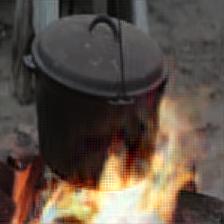}	
\includegraphics[width=.15\linewidth]{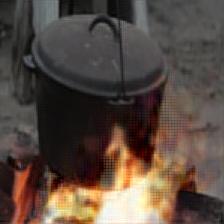}
\includegraphics[width=.15\linewidth]{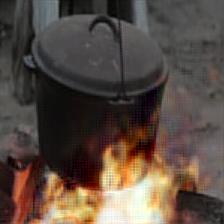}\\
(c) burning fire heating a pot \\[3px]

\includegraphics[width=.15\linewidth]{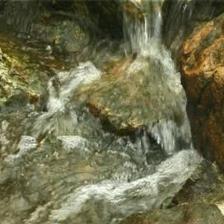}
\includegraphics[width=.15\linewidth]{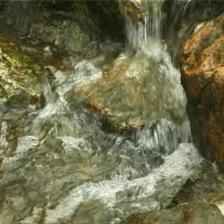}
\includegraphics[width=.15\linewidth]{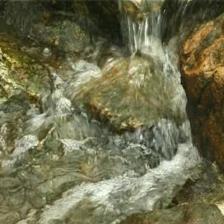}
\includegraphics[width=.15\linewidth]{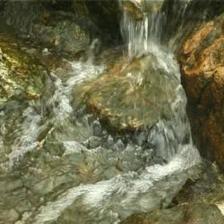}
\includegraphics[width=.15\linewidth]{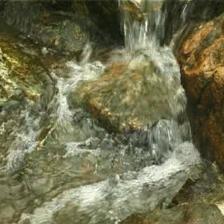}
\includegraphics[width=.15\linewidth]{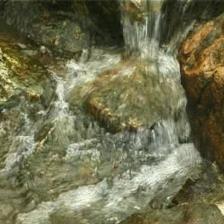} \\[3px]

\includegraphics[width=.15\linewidth]{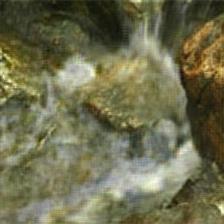}
\includegraphics[width=.15\linewidth]{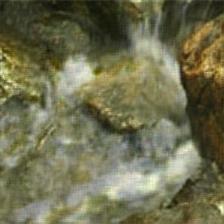}
\includegraphics[width=.15\linewidth]{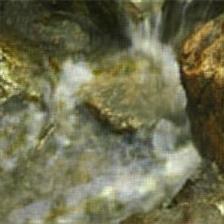}	
\includegraphics[width=.15\linewidth]{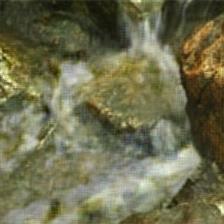}
\includegraphics[width=.15\linewidth]{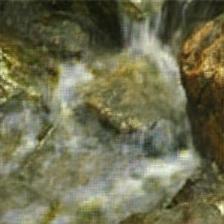}
\includegraphics[width=.15\linewidth]{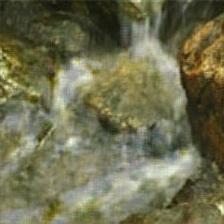}\\(d) spring water \\[3px]
	\caption{Synthesizing dynamic textures with only temporal stationarity. For each category, the first row displays the frames of the observed sequence, and the second row displays the corresponding frames of a synthesized sequence generated by the learning algorithm. (a) flashing lights. (b) fountain. (c) burning fire heating a pot. (d) spring water.}
	\label{fig:DTresults}
\end{center}
\end{figure}

\begin{figure}
\begin{center}
\includegraphics[width=.15\linewidth]{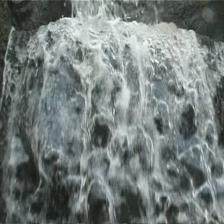}
\includegraphics[width=.15\linewidth]{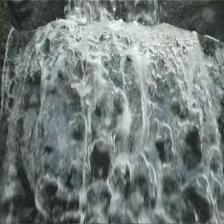}
\includegraphics[width=.15\linewidth]{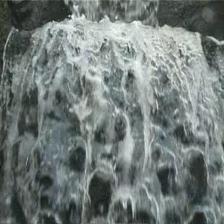}
\includegraphics[width=.15\linewidth]{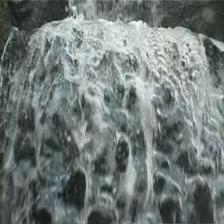}
\includegraphics[width=.15\linewidth]{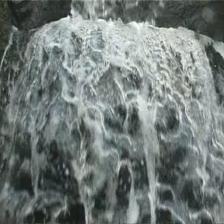}
\includegraphics[width=.15\linewidth]{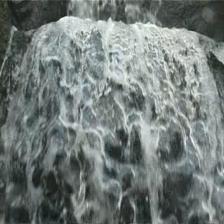} \\[3px]	

\includegraphics[width=.15\linewidth]{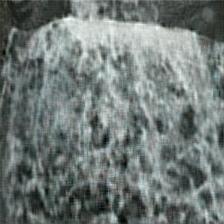}
\includegraphics[width=.15\linewidth]{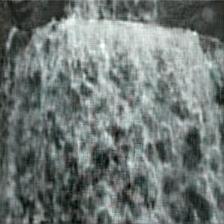}
\includegraphics[width=.15\linewidth]{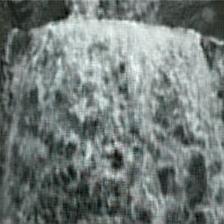}
\includegraphics[width=.15\linewidth]{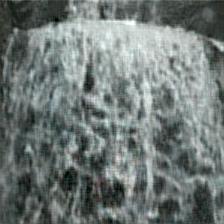}	
\includegraphics[width=.15\linewidth]{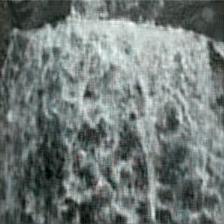}
\includegraphics[width=.15\linewidth]{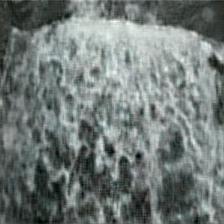}	\\[3px]

\includegraphics[width=.15\linewidth]{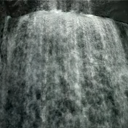}
\includegraphics[width=.15\linewidth]{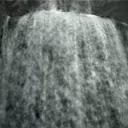}
\includegraphics[width=.15\linewidth]{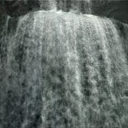}
\includegraphics[width=.15\linewidth]{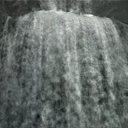}
\includegraphics[width=.15\linewidth]{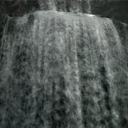}
\includegraphics[width=.15\linewidth]{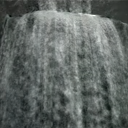} \\[3px]
	\caption{Comparison on synthesizing dynamic texture of waterfall. From top to bottom: segments of the observed sequence, synthesized sequence by
our method, and synthesized sequence by the
method of \cite{doretto2003dynamic}.}
	\label{fig:DT_comp}
\end{center}
\end{figure}

We first learn the model from dynamic textures that are stationary in both spatial and temporal domains. We use spatial-temporal filters that are convolutional in both spatial and temporal domains. The first layer has 120 $15 \times 15 \times 15$ filters with sub-sampling size of 7 pixels and frames. The second layer has 40 $7 \times 7 \times 7$ filters with sub-sampling size of 3.  The third layer has 20 $3 \times 3 \times 2$ filters with sub-sampling size of $2 \times 2 \times 1$. Figure \ref{fig:DTresults1} displays 2 results. For each category, the first row displays 7 frames of the observed sequence, while the second and third rows  show the corresponding frames of two synthesized sequences generated by the learning algorithm. 

We use the layer-by-layer learning scheme. Starting from the first layer, we sequentially add the layers one by one. Each time we learn the model and generate the synthesized image sequence using Algorithm \ref{code:FRAME}. While learning the new layer of filters, we refine the lower layers of filters with back-propagation.    


We learn a spatial-temporal generative ConvNet for each category from one observed video that is prepared to be of the size 224 $\times$ 224 $\times$ 50 or 70. The range of intensities is  [0, 255]. Mean subtraction is used as pre-processing. We use $\tilde{M}=3$ chain for Langevin sampling. The number of Langevin iterations between every two consecutive updates of parameters, $l = 20$. The number of learning iterations $T = 1200$, where we add one more layer every $400$ iterations.  We use layer-specific learning rates, where the learning rate at the higher layer is less than that at the lower layer, in order to obtain stable convergence.

\subsection{Experiment 2: Generating dynamic textures with only temporal stationarity}

Many dynamic textures have structured background and objects that are not stationary in the spatial domain. In this case, the network used in Experiment 1 may fail. However, we can modify the  network in Experiment 1  by using  filters that are fully connected in the spatial domain at the second layer. Specifically, the first layer has 120 $7 \times 7 \times 7$ filters with sub-sampling size of 3 pixels and frames. The second layer is a spatially fully connected layer, which contains 30 filters that are fully connected in the spatial domain but convolutional in the temporal domain. The temporal size of the filters is 4 frames with sub-sampling size of 2 frames in the temporal dimension.  Due to the spatial full connectivity at the second layer, the spatial domain of the feature maps at the third layer is reduced to $1 \times 1$.  The third layer has 5 $1 \times 1 \times 2$ filters with sub-sampling size of 1 in the temporal dimension. 

We  use  end-to-end learning scheme to learn the above 3-layer spatial-temporal generative ConvNet for dynamic textures. At each iteration, the 3 layers of filters are updated with 3 different layer-specific learning rates. The learning rate at the higher layer is much less than that at the lower layer to avoid the issue of large gradients. 

We learn a spatial-temporal generative ConvNet for each category from one training video. We synthesize $\tilde{M}=3$ videos using the Langevin dynamics. Figure \ref{fig:DTresults} displays the results. For each category, the first row shows 6 frames of the observed sequence (224 $\times$ 224 $\times$ 70), and the second  row shows the corresponding frames of a synthesized sequence generated by the learning algorithm. We use the same set of parameters for all the categories without tuning. Figure \ref{fig:DT_comp} compares our method to that of \cite{doretto2003dynamic}, which is a linear dynamic system model. The image sequence generated by this model appears more blurred than the sequence generated by our method.

\begin{figure}
\begin{center}
\includegraphics[width=.09\linewidth]{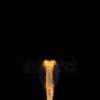}
\includegraphics[width=.09\linewidth]{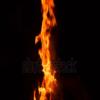}
\includegraphics[width=.09\linewidth]{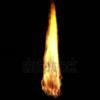}
\includegraphics[width=.09\linewidth]{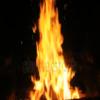}
\includegraphics[width=.09\linewidth]{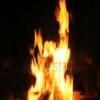}
\includegraphics[width=.09\linewidth]{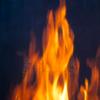}
\includegraphics[width=.09\linewidth]{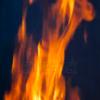}
\includegraphics[width=.09\linewidth]{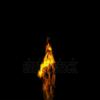}
\includegraphics[width=.09\linewidth]{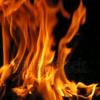}
\includegraphics[width=.09\linewidth]{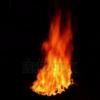} \\ \vspace{2pt}
\includegraphics[width=.09\linewidth]{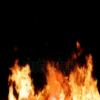}
\includegraphics[width=.09\linewidth]{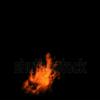}
\includegraphics[width=.09\linewidth]{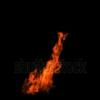}
\includegraphics[width=.09\linewidth]{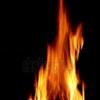}
\includegraphics[width=.09\linewidth]{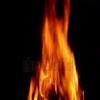}
\includegraphics[width=.09\linewidth]{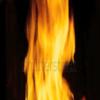}
\includegraphics[width=.09\linewidth]{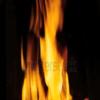}
\includegraphics[width=.09\linewidth]{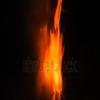}
\includegraphics[width=.09\linewidth]{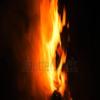}
\includegraphics[width=.09\linewidth]{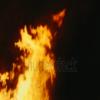} \\ \vspace{2pt}
\includegraphics[width=.09\linewidth]{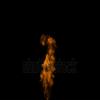}
\includegraphics[width=.09\linewidth]{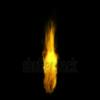}
\includegraphics[width=.09\linewidth]{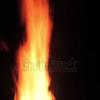}
\includegraphics[width=.09\linewidth]{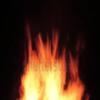}
\includegraphics[width=.09\linewidth]{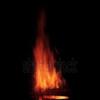}
\includegraphics[width=.09\linewidth]{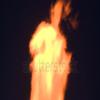}
\includegraphics[width=.09\linewidth]{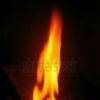}
\includegraphics[width=.09\linewidth]{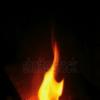}
\includegraphics[width=.09\linewidth]{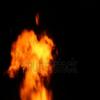}
\includegraphics[width=.09\linewidth]{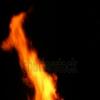}\\  \vspace{-3pt} {\footnotesize (a) $21$-st frame of 30 observed sequences} \\ \vspace{3pt}

\includegraphics[width=.09\linewidth]{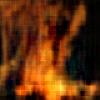}
\includegraphics[width=.09\linewidth]{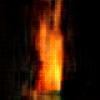}
\includegraphics[width=.09\linewidth]{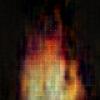}
\includegraphics[width=.09\linewidth]{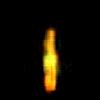}
\includegraphics[width=.09\linewidth]{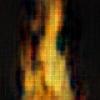}
\includegraphics[width=.09\linewidth]{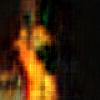}
\includegraphics[width=.09\linewidth]{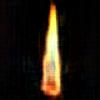}
\includegraphics[width=.09\linewidth]{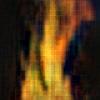}
\includegraphics[width=.09\linewidth]{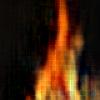}
\includegraphics[width=.09\linewidth]{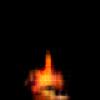}\\ \vspace{2pt}
\includegraphics[width=.09\linewidth]{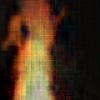}
\includegraphics[width=.09\linewidth]{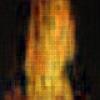}
\includegraphics[width=.09\linewidth]{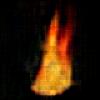}
\includegraphics[width=.09\linewidth]{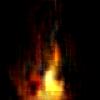}
\includegraphics[width=.09\linewidth]{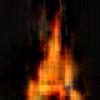}
\includegraphics[width=.09\linewidth]{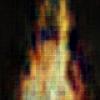}
\includegraphics[width=.09\linewidth]{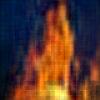}
\includegraphics[width=.09\linewidth]{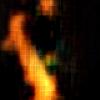}
\includegraphics[width=.09\linewidth]{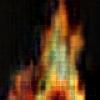}
\includegraphics[width=.09\linewidth]{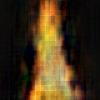}\\ \vspace{2pt}
\includegraphics[width=.09\linewidth]{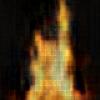}
\includegraphics[width=.09\linewidth]{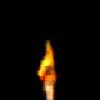}
\includegraphics[width=.09\linewidth]{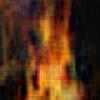}
\includegraphics[width=.09\linewidth]{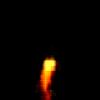}
\includegraphics[width=.09\linewidth]{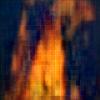}
\includegraphics[width=.09\linewidth]{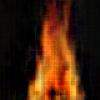}
\includegraphics[width=.09\linewidth]{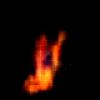}
\includegraphics[width=.09\linewidth]{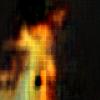}
\includegraphics[width=.09\linewidth]{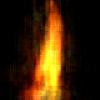}
\includegraphics[width=.09\linewidth]{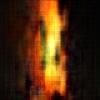}\\  \vspace{-3pt} {\footnotesize (b) $21$-st frame of 30 synthesized sequences} \\ \vspace{3pt}

\includegraphics[width=.09\linewidth]{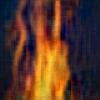}
\includegraphics[width=.09\linewidth]{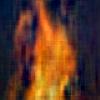}
\includegraphics[width=.09\linewidth]{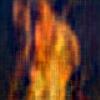}
\includegraphics[width=.09\linewidth]{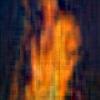}
\includegraphics[width=.09\linewidth]{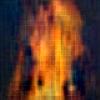}
\includegraphics[width=.09\linewidth]{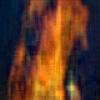}
\includegraphics[width=.09\linewidth]{./big_data/iter_1350/synthesis_32/image_0021.jpg}
\includegraphics[width=.09\linewidth]{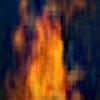}
\includegraphics[width=.09\linewidth]{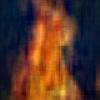}
\includegraphics[width=.09\linewidth]{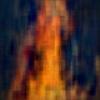}\\ \vspace{3pt}

\includegraphics[width=.09\linewidth]{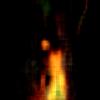}
\includegraphics[width=.09\linewidth]{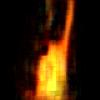}
\includegraphics[width=.09\linewidth]{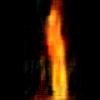}
\includegraphics[width=.09\linewidth]{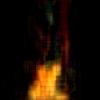}
\includegraphics[width=.09\linewidth]{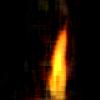}
\includegraphics[width=.09\linewidth]{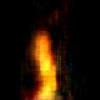}
\includegraphics[width=.09\linewidth]{./big_data/iter_1350/synthesis_38/image_0021.jpg}
\includegraphics[width=.09\linewidth]{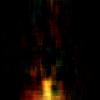}
\includegraphics[width=.09\linewidth]{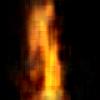}
\includegraphics[width=.09\linewidth]{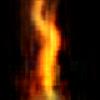}\\  \vspace{-3pt} {\footnotesize (c) 2 examples of synthesized sequences} \\ \vspace{3pt}
	\caption{Learning  from 30 observed fire videos with mini-batch implementation.}
	\label{fig:flames}
\end{center}
\end{figure}

The learning of our model can be scaled up. We learn the fire pattern from 30 training videos, with mini-batch implementation. The size of each mini-batch is 10 videos. Each video contains 30 frames ($100 \times 100$ pixels). For each mini-batch, ${\tilde M} = 13$ parallel chains for Langevin sampling is used. For this experiment, we slightly modify the network by using $120$ $11 \times 11 \times 9$ filters with sub-sampling size of $5$ pixels and $4$ frames at the first layer, and 30 spatially fully connected filters with temporal size of 5 frames and sub-sampling size of $2$ at the second layer, while keeping the setting of the third layer unchanged. The number of learning iterations $T=1300$. Figure \ref{fig:flames} shows one frame for each of 30 observed sequences and the corresponding frame of the synthesized sequences. Two examples of synthesized sequences are also displayed.

\subsection{Experiment 3: Generating action patterns without spatial or temporal stationarity}

Experiments 1 and 2 show  that the generative spatial-temporal ConvNet can learn from sequences without alignment. We can also specialize it to learning roughly aligned video sequences of action patterns, which are non-stationary in either spatial or temporal domain, by using a single
top-layer filter that covers the whole video sequence. We learn a 2-layer spatial-temporal generative  ConvNet from video sequences of aligned actions. The first layer has 200 $7 \times 7  \times 7$ filters with sub-sampling size of 3 pixels and frames. The second layer is a fully connected layer with a single filter that covers the whole sequence. The observed sequences are of the size $100 \times 200 \times 70$.

\begin{figure}
\begin{center}
\includegraphics[width=.15\linewidth]{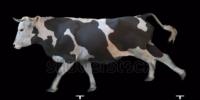}
\includegraphics[width=.15\linewidth]{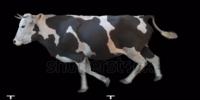}
\includegraphics[width=.15\linewidth]{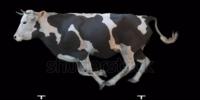}
\includegraphics[width=.15\linewidth]{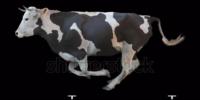}
\includegraphics[width=.15\linewidth]{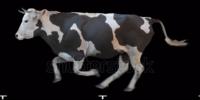}
\includegraphics[width=.15\linewidth]{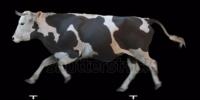} \\ \vspace{1pt}

\includegraphics[width=.15\linewidth]{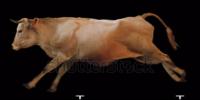}
\includegraphics[width=.15\linewidth]{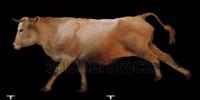}
\includegraphics[width=.15\linewidth]{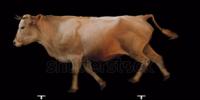}
\includegraphics[width=.15\linewidth]{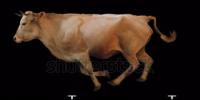}
\includegraphics[width=.15\linewidth]{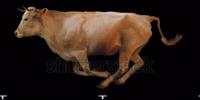}
\includegraphics[width=.15\linewidth]{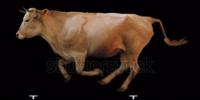}
\\ \vspace{-3pt} {\footnotesize observed sequences} \\ \vspace{3pt}

\includegraphics[width=.15\linewidth]{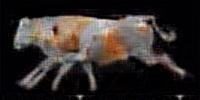}
\includegraphics[width=.15\linewidth]{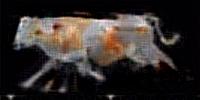}
\includegraphics[width=.15\linewidth]{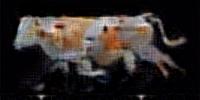}
\includegraphics[width=.15\linewidth]{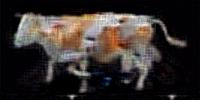}
\includegraphics[width=.15\linewidth]{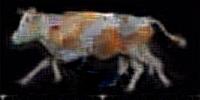}
\includegraphics[width=.15\linewidth]{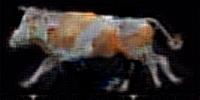} \\ \vspace{1pt}

\includegraphics[width=.15\linewidth]{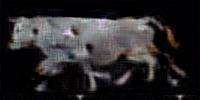}
\includegraphics[width=.15\linewidth]{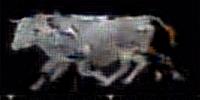}
\includegraphics[width=.15\linewidth]{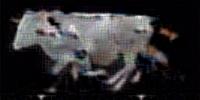}
\includegraphics[width=.15\linewidth]{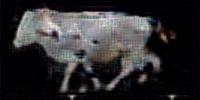}
\includegraphics[width=.15\linewidth]{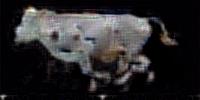}
\includegraphics[width=.15\linewidth]{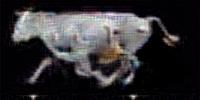}
\\ \vspace{-3pt} {\footnotesize synthesized sequences} \\ \vspace{2pt} (a) running cows \vspace{5pt}\\

\includegraphics[width=.15\linewidth]{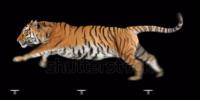}
\includegraphics[width=.15\linewidth]{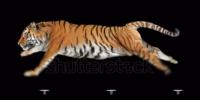}
\includegraphics[width=.15\linewidth]{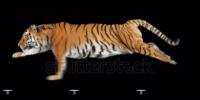}
\includegraphics[width=.15\linewidth]{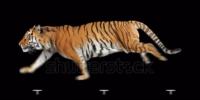}
\includegraphics[width=.15\linewidth]{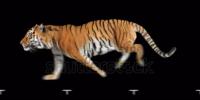}
\includegraphics[width=.15\linewidth]{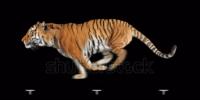} \\ \vspace{1pt}

\includegraphics[width=.15\linewidth]{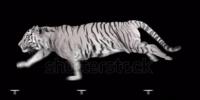}
\includegraphics[width=.15\linewidth]{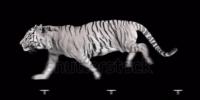}
\includegraphics[width=.15\linewidth]{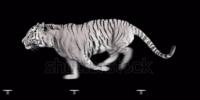}
\includegraphics[width=.15\linewidth]{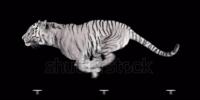}
\includegraphics[width=.15\linewidth]{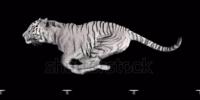}
\includegraphics[width=.15\linewidth]{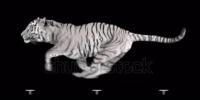}\\  \vspace{-3pt} {\footnotesize observed sequences} \vspace{3pt}

\includegraphics[width=.15\linewidth]{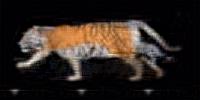}
\includegraphics[width=.15\linewidth]{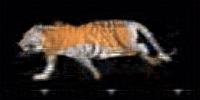}
\includegraphics[width=.15\linewidth]{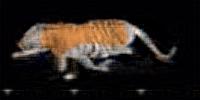}
\includegraphics[width=.15\linewidth]{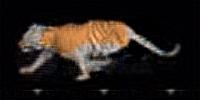}
\includegraphics[width=.15\linewidth]{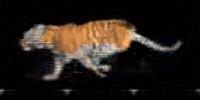}
\includegraphics[width=.15\linewidth]{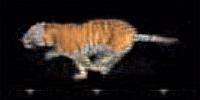} \\ \vspace{1pt}

\includegraphics[width=.15\linewidth]{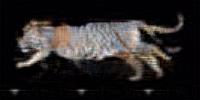}
\includegraphics[width=.15\linewidth]{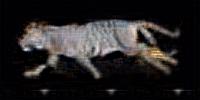}
\includegraphics[width=.15\linewidth]{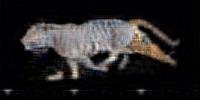}
\includegraphics[width=.15\linewidth]{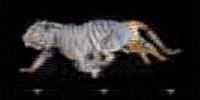}
\includegraphics[width=.15\linewidth]{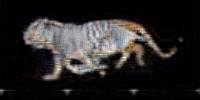}
\includegraphics[width=.15\linewidth]{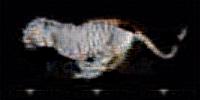}
\\  \vspace{-3pt} {\footnotesize synthesized sequences} \\  \vspace{2pt}(b) running tigers \\[3pt]
	\caption{Synthesizing action patterns. For each action video sequence, 6 continuous frames are shown. (a) running cows. Frames of 2 of 5 training sequences are displayed. The corresponding frames of 2 of 8 synthesized sequences generated by the learning algorithm are displayed. (b) running tigers. Frames of 2 observed training sequences are displayed. The corresponding frames of 2 of 4 synthesized sequences are displayed.}
	\label{fig:action}
\end{center}
\end{figure}

Figure \ref{fig:action} displays two results of modeling and synthesizing actions from roughly  aligned video sequences. We learn a model for each category, where the number of training sequences  is 5 for the running cow example, and 2 for the running tiger example. The videos are collected from the Internet and each has 70 frames. For each example, Figure \ref{fig:action} displays segments of 2 observed sequences, and segments of 2 synthesized action sequences generated by the learning algorithm. We run $\tilde{M}=8$ paralleled  chains for the experiment of running cows, and 4 paralleled chains for the experiment of running tigers. The experiments show that our model can capture non-stationary action patterns.  

One limitation of our model is that it does not involve explicit tracking  of the objects and their parts. 
	  
\subsection{Experiment 4: Learning from incomplete data}

Our model can learn from video sequences with occluded pixels. The task is inspired by the fact that most of the videos contain occluded objects. Our learning method can be adapted to this task with minimal modification. The modification involves, for each iteration, running $k$ steps of Langevin dynamics to recover the occluded regions of the observed sequences. At each iteration, we use the  completed observed sequences and the synthesized sequences to compute the gradient of the log-likelihood and update the model parameters. Our method simultaneously accomplishes the following tasks: (1) recover the occluded pixels of the training video sequences, (2) synthesize new video sequences from the learned model, (3) learn the model by updating the model parameters using the recovered sequences and the synthesized sequences. See Algorithm \ref{code:recovery} for the  description of the learning, sampling, and recovery algorithm. 

\begin{algorithm}
\caption{Learning, sampling, and recovery algorithm}
\label{code:recovery}
\begin{algorithmic}[1]

\REQUIRE ~~\\
(1) training image sequences with occluded pixels $\{\I_m, m=1,...,M\}$ \\
(2) binary masks $\{O_m, m=1,...,M \}$ indicating the locations of the occluded pixels in the training image sequences\\  
(3) number of synthesized image sequences $\tilde{M}$\\
(4) number of Langevin steps $l$ for synthesizing image sequences\\
(5) number of Langevin steps $k$ for recovering the occluded pixels\\
(6) number of learning iterations $T$

\ENSURE~~\\
(1) estimated parameters $w$\\
(2) synthesized image sequences $\{\tI_m, m = 1, ..., \tilde{M}\}$ 
(3) recovered image sequences $\{\I^{'}_m, m=1,...,M\}$
\item[]
\STATE Let $t\leftarrow 0$, initialize $w^{(0)}$.
\STATE Initialize $\tI_m $, for $m = 1, ..., \tilde{M}$. 
\STATE Initialize $\I^{'}_m$, for $m = 1, ..., M$. 
\REPEAT 
\STATE For each $m$, run $k$ steps of Langevin dynamics to recover the occluded region of $\I^{'}_m$, i.e., starting from the current $\I^{'}_m$, each step 
follows equation (\ref{eq:Langevin}), but only the occluded pixels in $\I^{'}_m$ are updated in each step. 
\STATE For each $m$, run $l$ steps of Langevin dynamics to update $\tI_m$, i.e., starting from the current $\tI_m$, each step 
follows equation (\ref{eq:Langevin}). 
\STATE Calculate  $H^{\obs} = \sum_{m=1}^{M} \frac{\partial}{\partial w} f(\I^{'}_m; w^{(t)})/M$, and
$H^{\syn} =  \sum_{m=1}^{\tM} \frac{\partial}{\partial w} f(\tI_m; w^{(t)})/\tM$.
\STATE Update $w^{(t+1)} \leftarrow w^{(t)} + \eta ( H^{\obs} - H^{\syn}) $,  with step size $\eta$. 
\STATE Let $t \leftarrow t+1$
\UNTIL $t = T$
\end{algorithmic}
\end{algorithm}

\begin{table}[h]
\caption{Recovery errors in occlusion experiments}\label{recoveryExp}
\vskip 0.02in
\begin{center}
\begin{footnotesize}
(a) salt and pepper masks\\
\begin{tabular}{|c|c|c|c|}
\hline
           & ours   & MRF-$\ell_1$      & MRF-$\ell_2$  \\ \hline \hline
flag       & \textbf{3.7923} & 6.6211  & 10.9216  \\ \hline
fountain   & \textbf{5.5403} & 8.1904  & 11.3850  \\ \hline
ocean      & \textbf{3.3739} & 7.2983 & 9.6020     \\ \hline
playing    & \textbf{5.9035} & 14.3665& 15.7735    \\ \hline
sea world & \textbf{5.3720} &  10.6127  & 11.7803  \\ \hline
traffic    & \textbf{7.2029} & 14.7512 & 17.6790  \\ \hline
windmill   & \textbf{5.9484} &  8.9095   & 12.6487  \\ \hline \hline
Avg.   & \textbf{5.3048} &  10.1071    & 12.8272  \\ \hline
\end{tabular}
\vskip 0.08in
(b) single region masks \\
\begin{tabular}{|c|c|c|c|}
\hline
           & ours    & MRF-$\ell_1$      & MRF-$\ell_2$  \\ \hline \hline
flag       & \textbf{8.1636}  & 10.6586  & 12.5300\\ \hline
fountain   & \textbf{6.0323}  & 11.8299  & 12.1696\\ \hline
ocean      & \textbf{3.4842}  & 8.7498 & 9.8078\\ \hline
playing    & \textbf{6.1575}  & 15.6296& 15.7085\\ \hline
sea world & \textbf{5.8850} & 12.0297  & 12.2868\\ \hline
traffic    & \textbf{6.8306}  & 15.3660 & 16.5787\\ \hline
windmill   & \textbf{7.8858} & 11.7355   & 13.2036\\ \hline \hline
Avg.   & \textbf{ 6.3484} & 12.2856    & 13.1836\\ \hline
\end{tabular}
\vskip 0.08in
(c) 50$\%$ missing frames\\
\begin{tabular}{|c|c|c|c|}
\hline
           & ours    & MRF-$\ell_1$      & MRF-$\ell_2$  \\ \hline \hline
flag       & \textbf{5.5992} & 10.7171  & 12.6317\\ \hline
fountain   & \textbf{8.0531} & 19.4331  & 13.2251\\ \hline
ocean      & \textbf{4.0428}  & 9.0838 & 9.8913\\ \hline
playing    & \textbf{7.6103 } & 22.2827& 17.5692\\ \hline
sea world & \textbf{5.4348} & 13.5101  &12.9305\\ \hline
traffic    & \textbf{8.8245}  & 16.6965 & 17.1830\\ \hline
windmill   & \textbf{7.5346} & 13.3364  & 12.9911 \\ \hline \hline
Avg.   & \textbf{ 6.7285} & 15.0085   & 13.7746\\ \hline
\end{tabular}
\end{footnotesize}
\end{center}
\end{table}

\begin{figure}[h]
\begin{center}

\includegraphics[width=.15\linewidth]{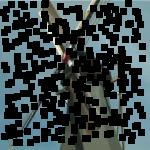}
\includegraphics[width=.15\linewidth]{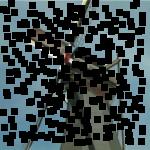}
\includegraphics[width=.15\linewidth]{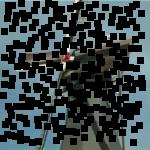}
\includegraphics[width=.15\linewidth]{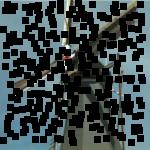}
\includegraphics[width=.15\linewidth]{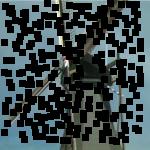}
\includegraphics[width=.15\linewidth]{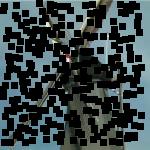} 
\\[3px]
\includegraphics[width=.15\linewidth]{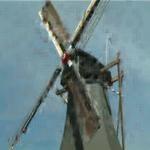}
\includegraphics[width=.15\linewidth]{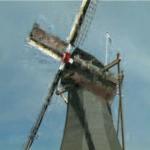}
\includegraphics[width=.15\linewidth]{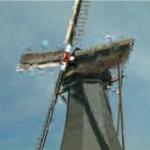}
\includegraphics[width=.15\linewidth]{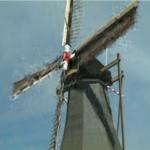}
\includegraphics[width=.15\linewidth]{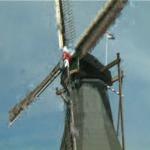}
\includegraphics[width=.15\linewidth]{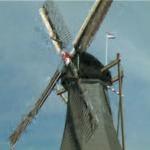}
\\ (a) $50\%$ salt and pepper masks \\[3px]

\includegraphics[width=.15\linewidth]{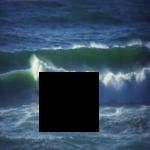}
\includegraphics[width=.15\linewidth]{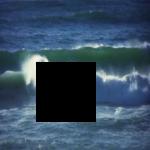}
\includegraphics[width=.15\linewidth]{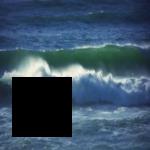}
\includegraphics[width=.15\linewidth]{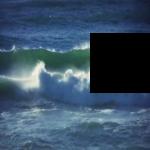}
\includegraphics[width=.15\linewidth]{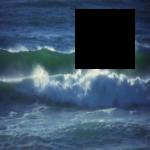}
\includegraphics[width=.15\linewidth]{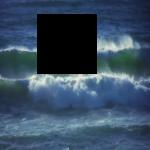} \\ [3px]
\includegraphics[width=.15\linewidth]{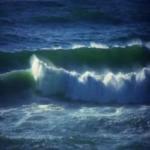}
\includegraphics[width=.15\linewidth]{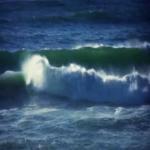}
\includegraphics[width=.15\linewidth]{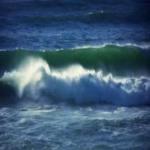}
\includegraphics[width=.15\linewidth]{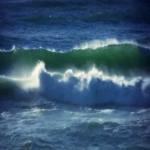}
\includegraphics[width=.15\linewidth]{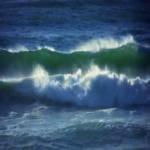}
\includegraphics[width=.15\linewidth]{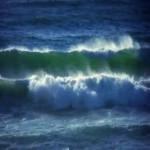}
\\ (b) single region masks \\[3px]


\includegraphics[width=.15\linewidth]{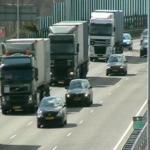}
\includegraphics[width=.15\linewidth]{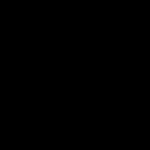}
\includegraphics[width=.15\linewidth]{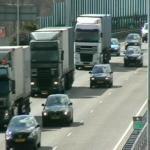}
\includegraphics[width=.15\linewidth]{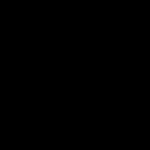}
\includegraphics[width=.15\linewidth]{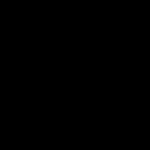}
\includegraphics[width=.15\linewidth]{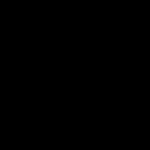}
\\[3px]
\includegraphics[width=.15\linewidth]{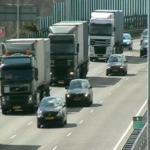}
\includegraphics[width=.15\linewidth]{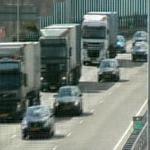}
\includegraphics[width=.15\linewidth]{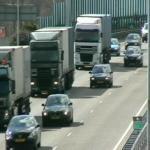}
\includegraphics[width=.15\linewidth]{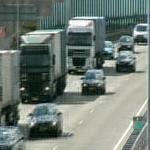}
\includegraphics[width=.15\linewidth]{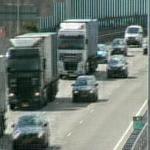}
\includegraphics[width=.15\linewidth]{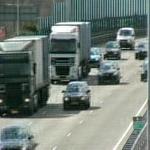}
\\ (c) $50\%$ missing frames \vspace{3pt}
	\caption{Learning from occluded video sequences. For each experiment, the first row shows a segment of the occluded sequence with black masks. The second row shows the corresponding segment of the recovered sequence. (a) type 1: salt and pepper mask.  (b) type 2: single region mask. (c) type 3: missing frames.}
	\label{fig:recovery}
\end{center}
\end{figure}

  

\begin{figure}[h]
\begin{center}
%
%
%
%
%
%
%
%

\includegraphics[width=.27\linewidth]{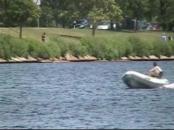}
\includegraphics[width=.27\linewidth]{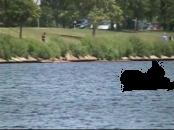}
\includegraphics[width=.27\linewidth]{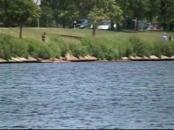}\\[3px]

\includegraphics[width=.27\linewidth]{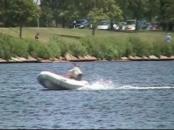}
\includegraphics[width=.27\linewidth]{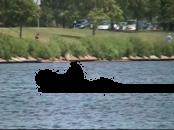}
\includegraphics[width=.27\linewidth]{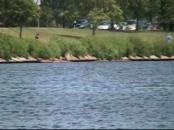}\\
(a) removing a moving boat in the lake\\[3px]

\includegraphics[width=.27\linewidth]{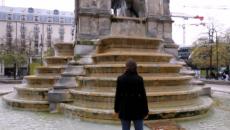}
\includegraphics[width=.27\linewidth]{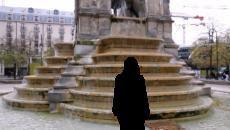}
\includegraphics[width=.27\linewidth]{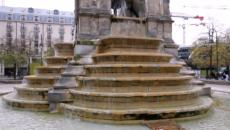}\\[3px]

\includegraphics[width=.27\linewidth]{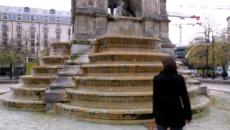}
\includegraphics[width=.27\linewidth]{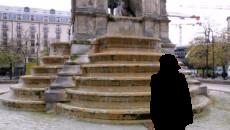}
\includegraphics[width=.27\linewidth]{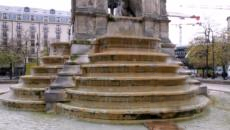}\\
(b) removing a walking person in front of fountain\\[3px]
	\caption{Background inpainting for videos. For each experiment, the first column displays 2 frames of the original video. The second column shows the corresponding frames with black masks occluding the target to be removed. The third column shows the inpainting result by our algorithm. (a) moving boat. (b) walking person.}
	\label{fig:bg_inpainting}
\end{center}
\end{figure}

We design 3 types of occlusions: (1) Type 1: salt and pepper occlusion, where we randomly place $7 \times 7$ masks on the $150 \times 150$ image domain to cover $50\%$ of the pixels of the videos. (2) Type 2: single region mask occlusion, where we randomly place a $60 \times 60$ mask on the $150 \times 150$ image domain. (3) Type 3: missing frames, where we randomly block $50\%$ of the image frames from each video. Figure \ref{fig:recovery} displays one example of the recovery result for each type of occlusion. Each video has 70 frames. 

To quantitatively evaluate the qualities of the recovered videos, we test our method on 7 video sequences, which are collected from DynTex++ dataset of \cite{ghanem2010maximum}, with 3 types of occlusions. We use the same model structure as the one used in Experiment 3. The number of Langevin steps for recovering is set to be equal to the number of Langevin steps for synthesizing, which is 20. For each experiment, we report the recovery errors measured by the average per pixel difference between the original image sequence and the recovered image sequence on the occluded pixels. The range of pixel intensities is $[0,255]$. We compare our results with the results obtained by a generic Markov random field model defined on the video sequence. The model is a 3D (spatial-temporal) Markov random field, whose potentials are pairwise $\ell_1$ or $\ell_2$ differences between nearest neighbor pixels, where the nearest neighbors are defined in both the spatial and temporal domains. The image sequences are recovered by sampling the intensities of the occluded pixels conditional on the observed pixels using the Gibbs sampler. Table \ref{recoveryExp} shows the comparison results for 3 types of occlusions. We can see that our model can recover the incomplete data, while learning from them.

\subsection{Experiment 5: Background inpainting}

If a moving object in the video is occluded in each frame, it turns out that the recovery algorithm will become an algorithm for background inpainting of videos, where the goal is to remove the undesired moving object from the video. We use the same model as the one in Experiment 2  for Figure \ref{fig:DTresults}. Figure \ref{fig:bg_inpainting} shows two examples of removals of (a) a moving boat and (b) a walking person  respectively. The videos are collected from \cite{data}. For each example, the first column displays 2 frames of the original video. The second column shows the corresponding frames with masks occluding the target to be removed. The third column presents the inpainting result by our algorithm. The video size is $130 \times 174 \times 150$ in example (a) and $130 \times 230 \times 104$ in example (b). The experiment is different from the video inpainting by interpolation. We synthesize image patches to fill in the empty regions of the video by running Langevin dynamics. For both Experiments 4 and 5, we run a single Langevin chain for synthesis.

\section{Conclusion}

In this paper, we propose a spatial-temporal generative ConvNet model for synthesizing dynamic patterns, such as dynamic textures and action patterns. Our experiments show that the model can synthesize realistic dynamic patterns. 
Moreover, it is possible to learn the model from video sequences with occluded pixels or missing frames.  

Other experiments, not included in this paper, show that our method can also generate sound patterns. 

The MCMC sampling of  the model can be sped up by learning and sampling the models at multiple scales, or by recruiting the generator network to reconstruct and regenerate the synthesized examples as in cooperative training \cite{xie2016cooperative}. 

\section*{Acknowledgments}
{\noindent{The work is supported by NSF DMS 1310391, DARPA SIMPLEX N66001-15-C-4035,  ONR MURI N00014-16-1-2007, and  DARPA ARO W911NF-16-1-0579. }}


{\small
\bibliographystyle{ieee}
\bibliography{mybibfile}

\begin{thebibliography}{10}\itemsep=-1pt

\bibitem{Denton2015a}
E.~L. Denton, S.~Chintala, R.~Fergus, et~al.
\newblock Deep generative image models using a laplacian pyramid of adversarial
  networks.
\newblock In {\em NIPS}, pages 1486--1494, 2015.

\bibitem{doretto2003dynamic}
G.~Doretto, A.~Chiuso, Y.~N. Wu, and S.~Soatto.
\newblock Dynamic textures.
\newblock {\em International Journal of Computer Vision}, 51(2):91--109, 2003.

\bibitem{Alexey2015}
A.~Dosovitskiy, J.~Tobias~Springenberg, and T.~Brox.
\newblock Learning to generate chairs with convolutional neural networks.
\newblock In {\em CVPR}, pages 1538--1546, 2015.

\bibitem{ghanem2010maximum}
B.~Ghanem and N.~Ahuja.
\newblock Maximum margin distance learning for dynamic texture recognition.
\newblock In {\em ECCV}, pages 223--236. Springer, 2010.

\bibitem{girolami2011riemann}
M.~Girolami and B.~Calderhead.
\newblock Riemann manifold langevin and hamiltonian monte carlo methods.
\newblock {\em Journal of the Royal Statistical Society: Series B (Statistical
  Methodology)}, 73(2):123--214, 2011.

\bibitem{goodfellow2014generative}
I.~Goodfellow, J.~Pouget-Abadie, M.~Mirza, B.~Xu, D.~Warde-Farley, S.~Ozair,
  A.~Courville, and Y.~Bengio.
\newblock Generative adversarial nets.
\newblock In {\em NIPS}, pages 2672--2680, 2014.

\bibitem{KarolICML2015}
K.~Gregor, I.~Danihelka, A.~Graves, D.~J. Rezende, and D.~Wierstra.
\newblock {DRAW:} {A} recurrent neural network for image generation.
\newblock In {\em ICML}, pages 1462--1471, 2015.

\bibitem{HanLu2016}
T.~Han, Y.~Lu, S.-C. Zhu, and Y.~N. Wu.
\newblock Alternating back-propagation for generator network.
\newblock In {\em AAAI}, 2017.

\bibitem{han2015video}
Z.~Han, Z.~Xu, and S.-C. Zhu.
\newblock Video primal sketch: A unified middle-level representation for video.
\newblock {\em Journal of Mathematical Imaging and Vision}, 53(2):151--170,
  2015.

\bibitem{hochreiter1997long}
S.~Hochreiter and J.~Schmidhuber.
\newblock Long short-term memory.
\newblock {\em Neural computation}, 9(8):1735--1780, 1997.

\bibitem{ji20133d}
S.~Ji, W.~Xu, M.~Yang, and K.~Yu.
\newblock 3d convolutional neural networks for human action recognition.
\newblock {\em IEEE Transactions on Pattern Analysis and Machine Intelligence},
  35(1):221--231, 2013.

\bibitem{krizhevsky2012imagenet}
A.~Krizhevsky, I.~Sutskever, and G.~E. Hinton.
\newblock Imagenet classification with deep convolutional neural networks.
\newblock In {\em NIPS}, pages 1097--1105, 2012.

\bibitem{Kulkarni2015}
T.~D. {Kulkarni}, W.~{Whitney}, P.~{Kohli}, and J.~B. {Tenenbaum}.
\newblock {Deep Convolutional Inverse Graphics Network}.
\newblock {\em ArXiv e-prints}, 2015.

\bibitem{lecun1998gradient}
Y.~LeCun, L.~Bottou, Y.~Bengio, and P.~Haffner.
\newblock Gradient-based learning applied to document recognition.
\newblock {\em Proceedings of the IEEE}, 86(11):2278--2324, 1998.

\bibitem{lecun2006tutorial}
Y.~LeCun, S.~Chopra, R.~Hadsell, M.~Ranzato, and F.~Huang.
\newblock A tutorial on energy-based learning.
\newblock {\em Predicting structured data}, 1:0, 2006.

\bibitem{LuZhuWu2016}
Y.~Lu, S.-C. Zhu, and Y.~N. Wu.
\newblock Learning {FRAME} models using cnn filters.
\newblock In {\em AAAI}, 2016.

\bibitem{montufar2014number}
G.~F. Montufar, R.~Pascanu, K.~Cho, and Y.~Bengio.
\newblock On the number of linear regions of deep neural networks.
\newblock In {\em NIPS}, pages 2924--2932, 2014.

\bibitem{neal2011mcmc}
R.~M. Neal.
\newblock Mcmc using hamiltonian dynamics.
\newblock {\em Handbook of Markov Chain Monte Carlo}, 2, 2011.

\bibitem{data}
A.~Newson, A.~Almansa, M.~Fradet, Y.~Gousseau, and P.~Pérez.
\newblock \url{http://perso.telecom-paristech.fr/~gousseau/video_inpainting}.

\bibitem{Ng2011}
J.~Ngiam, Z.~Chen, P.~W. Koh, and A.~Y. Ng.
\newblock Learning deep energy models.
\newblock In {\em ICML}, pages 1105--1112, 2011.

\bibitem{MexConv3D}
P.~Sun.
\newblock \url{https://github.com/pengsun/MexConv3D}.

\bibitem{matconvnn}
A.~Vedaldi and K.~Lenc.
\newblock Matconvnet -- convolutional neural networks for matlab.
\newblock {\em CoRR}, abs/1412.4564, 2014.

\bibitem{vondrick2016generating}
C.~Vondrick, H.~Pirsiavash, and A.~Torralba.
\newblock Generating videos with scene dynamics.
\newblock In {\em NIPS}, pages 613--621, 2016.

\bibitem{wang2002generative}
Y.~Wang and S.-C. Zhu.
\newblock A generative method for textured motion: Analysis and synthesis.
\newblock In {\em ECCV}, pages 583--598. Springer, 2002.

\bibitem{wang2004analysis}
Y.~Wang and S.-C. Zhu.
\newblock Analysis and synthesis of textured motion: Particles and waves.
\newblock {\em IEEE Transactions on Pattern Analysis and Machine Intelligence},
  26(10):1348--1363, 2004.

\bibitem{welling2009herding}
M.~Welling.
\newblock Herding dynamical weights to learn.
\newblock In {\em ICML}, pages 1121--1128. ACM, 2009.

\bibitem{williams1989learning}
R.~J. Williams and D.~Zipser.
\newblock A learning algorithm for continually running fully recurrent neural
  networks.
\newblock {\em Neural computation}, 1(2):270--280, 1989.

\bibitem{xie2016cooperative}
J.~Xie, Y.~Lu, R.~Gao, S.-C. Zhu, and Y.~N. Wu.
\newblock Cooperative training of descriptor and generator networks.
\newblock {\em arXiv preprint arXiv:1609.09408}, 2016.

\bibitem{XieLuICML}
J.~Xie, Y.~Lu, S.-C. Zhu, and Y.~N. Wu.
\newblock A theory of generative convnet.
\newblock In {\em ICML}, 2016.

\bibitem{you2016kernel}
X.~You, W.~Guo, S.~Yu, K.~Li, J.~C. Pr{\'\i}ncipe, and D.~Tao.
\newblock Kernel learning for dynamic texture synthesis.
\newblock {\em IEEE Transactions on Image Processing}, 25(10):4782--4795, 2016.

\bibitem{younes1999convergence}
L.~Younes.
\newblock On the convergence of markovian stochastic algorithms with rapidly
  decreasing ergodicity rates.
\newblock {\em Stochastics: An International Journal of Probability and
  Stochastic Processes}, 65(3-4):177--228, 1999.

\bibitem{zeiler2011adaptive}
M.~D. Zeiler, G.~W. Taylor, and R.~Fergus.
\newblock Adaptive deconvolutional networks for mid and high level feature
  learning.
\newblock In {\em ICCV}, pages 2018--2025, 2011.

\end{thebibliography}
}

\end{document}